\documentclass[10pt,journal,compsoc]{IEEEtran}

\usepackage{amsmath,bm, amsfonts,color,colordvi,caption,subcaption,url}
\usepackage{mathrsfs}
\usepackage{CJK}
\usepackage{makecell}
\usepackage{picinpar}
\usepackage{multirow}
\usepackage{picinpar}
\usepackage{algorithm}
\usepackage{algorithmic}
\usepackage{ragged2e}
\usepackage{array}
\usepackage[colorlinks,
            linkcolor=red,
            anchorcolor=green,
            citecolor=blue
            ]{hyperref}
\usepackage[noadjust]{cite}
\usepackage{amssymb}
\usepackage[subfigure]{tocloft}
\usepackage[dvips]{graphicx}
\usepackage{float}
\usepackage{subcaption}
\usepackage{booktabs}

\usepackage[T1]{fontenc}

\usepackage{lipsum}

\newtheorem{theorem}{Theorem}[section]

\def\R{{\mathbb R}}

\def\bfc{{\bf c}}

\def\bfx{{\bf x}}
\def\bfw{{\bf w}}

\def\bfX{{\bf X}}
\def\bfY{{\bf Y}}

\usepackage{hyperref}
\hypersetup{colorlinks,breaklinks,
            citecolor=[rgb]{1,0,0},
            linkcolor=[rgb]{0,0,1}}
\usepackage[nameinlink,noabbrev,capitalize]{cleveref}

\ifCLASSINFOpdf
\else
\fi

\ifCLASSOPTIONcompsoc
\else
  \usepackage{cite}
\fi

\begin{document}
\title{Generalization-Memorization Machines}

\author{ {Zhen Wang, Yuanhai Shao$^*$}
\IEEEcompsocitemizethanks{\IEEEcompsocthanksitem Z. Wang is with the School of Mathematical Sciences, Inner Mongolia University, Hohhot, 010021, P.R. China.
Email: wangzhen@imu.edu.cn.\protect
\IEEEcompsocthanksitem Y.H. Shao (*Corresponding author) is with the Management School, Hainan University, Haikou, P.R. China. Email: shaoyuanhai@hainanu.edu.cn.\protect
}

\thanks{Manuscript received xx, xx; revised xx, xx.}}

\IEEEtitleabstractindextext{\justify
\begin{abstract}
Classifying the training data correctly without over-fitting is one of the goals in machine learning. In this paper, we propose a generalization-memorization mechanism, including a generalization-memorization decision and a memory modeling principle. Under this mechanism, error-based learning machines improve their memorization abilities of training data without over-fitting. Specifically, the generalization-memorization machines (GMM) are proposed by applying this mechanism. The optimization problems in GMM are quadratic programming problems and could be solved efficiently. It should be noted that the recently proposed generalization-memorization kernel and the corresponding support vector machines are the special cases of our GMM. Experimental results show the effectiveness of the proposed GMM both on memorization and generalization.
\end{abstract}
\begin{IEEEkeywords}
Classification, support vector machines, generalization-memorization mechanism, memory machines, generalization-memorization kernel.
\end{IEEEkeywords}}

\maketitle

\IEEEdisplaynontitleabstractindextext

%
\IEEEpeerreviewmaketitle

\ifCLASSOPTIONcompsoc
\IEEEraisesectionheading{\section{Introduction}\label{sec:introduction}}
\else
\section{Introduction}
\label{sec:introduction}
\fi
\IEEEPARstart{M}{emory} is both important for humans and learning machines \cite{memorizationDNN2018,Memory1}. For a learning machine, the label memory represents the ability to memorize the training samples, also known as the ability of classifying the training samples correctly. Generally, classifying all of the training samples correctly is called zero empirical risk. Most of machine learning models pursue a well generalization performance by minimizing the empirical risks with regularization terms \cite{DeepL,MachineL}. However, the problem of over-fitting may appear if we pursue to minimize the empirical risk in machine learning, especially for small data problems. Recent studies \cite{memorizationDNN2017,memorizationDNN2018,memorizationDNN2020} show that the deep neural networks (DNN) obtain almost zero empirical risk with a well generalization performance. A natural question is whether all machine learning models could obtain almost zero empirical risks with well generalization performances.

Recently, Vapnik and Izmailov \cite{MSVM2021} studied the memorization problem of support vector machines (SVM) \cite{SVM95}, and a generalization-memorization kernel which is the sum of two RBF kernels \cite{Kernel1} is introduced into SVM to improve its memorization ability, called SVM$^{m}$. By adjusting the generalization-memorization kernel parameters properly, SVM$^m$ could classify training samples correctly while with a comparable generalization compared with SVM. Though \cite{MSVM2021} shows a path to realize both of the generalization and memorization of SVM, the performances of applying the generalization-memorization kernel into other SVM-type models remain unclear. More importantly, it is difficult to apply the generalization-memorization kernel into other non-kernel machine learning models \cite{DeepL,RandomF,Nonkernel1,Nonkernel2} by this path. Therefore, how to keep machine learning models have almost zero empirical risks with well generalization performances is still a problem.


%

In this paper, we propose a
generalization-memorization mechanism for machine learning models to improve the memorization ability with competitive generalizations. This mechanism contains a new generalization-memorization decision and a memory modeling principle.
The memorization ability of the new decision relies on a memory cost function, and it affects the generalization by a new memory influence function. In memory modeling, the principle gives a path to preserve the generalization. Further, we apply the generalization-memorization mechanism in SVM to propose the generalization-memorization machines (GMM), which can obtain zero empirical risks with competitive generalizations.

The main contributions of this paper are:

\noindent (i) For label memory problems, we propose a generalization-memorization mechanism to guide the generalization and memorization of the learning models. The proposed mechanism could be applied into various error-based machine learning models directly, whatever kernel or non-kernel based models.

\noindent (ii) Generalization-memorization machines (GMM) are proposed based on the above generalization-memorization mechanism, including a hard version (HGMM) and a soft version (SGMM). The HGMM requires to classify all of the training samples correctly, and the necessary and sufficient condition that HGMM has a zero empirical risk is given. In theory, a convenient sufficient condition is also given for practical applications. When the zero empirical risk is unnecessary, the HGMM is relaxed to the SGMM to obtain a competitive generalization ability. Particularly, our HGMM and SGMM are very simple and their problem sizes are the same as SVM.

\noindent (iii) The generalization-memorization decision endows the generalization-memorization kernel of SVM$^m$ \cite{MSVM2021} with more clear geometric interpretations, and it is proved that the SVM$^m$ is a special case of our SGMM.

\noindent (iv) Experimental results confirm the well performance of the proposed GMM on both memorization and generalization.

This paper is organized as follows. In the next section, a brief overview of SVM and SVM$^m$ is given. Section 3 establishes the new generalization-memorization mechanism, including a generalization-memorization decision and a memory modeling principle. Section 4 proposes the GMM and discusses the relationship between GMM and SVMs. Numerical experiments and concluding remarks are given in the last two sections.

\section{Review of SVM and SVM$^m$}
In this paper, we consider a binary classification problem in the $n$-dimensional real
space $\R^n$. The set of training samples is represented by
$T=\{(\bfx_{i},y_{i})|i=1,2,...,m\}$, where $\bfx_{i}\in \R^{n}$ is the input and $y_{i}\in\{+1,-1\}$ is the
corresponding label. The training samples together with their labels are organized in matrix $\bfX\in\R^{n\times m}$ and diagonal matrix $\bfY$ with diagonal elements $\bfY_{ii}=y_i$ ($i=1,\ldots,m$), respectively.

The classical SVM \cite{SVM95,SVMbook} constructs a classification decision $f(\bfx)=\langle \bfw, \varphi(\bfx) \rangle +b$ with $\bfw\in \R^d$ and $b\in\R$ by solving
\begin{equation}\label{SVMprimal}
\begin{array}{ll}
 \underset{\bfw,b,\xi_\cdot}{\min} &~\displaystyle \frac{1}{2}||\bfw||^2+ C \displaystyle \sum_{i=1}^{m}\displaystyle \xi_{i},\\
 \hbox{s.t.\ }&~y_{i}(\langle \bfw, \varphi(\bfx_{i}) \rangle +b)\geq 1-\xi_{i} , \xi_{i}\geq 0,
 i=1,\cdots,m,
\end{array}
\end{equation}
where $\langle\cdot,\cdot\rangle$ denotes the inner product, $\varphi(\cdot)$ is a mapping, $C$ is a positive parameter, and $\xi_i$ is a slack variable.
For a new sample $\bfx$, it would be classified into positive class with $y=+1$ if $\langle \bfw, \varphi(\bfx) \rangle +b>0$, and otherwise, it is classified into the negative class with $y=-1$.
Generally, we solve the dual problem of \eqref{SVMprimal} as
\begin{equation}\label{SVMdual}
\begin{array}{ll}
 \underset{\alpha}{\min}&\frac{1}{2}\alpha^\top\bfY K(\bfX,\bfX)\bfY\alpha-e^\top\alpha\\
\mbox{s.\,t.\,}&  e^\top\bfY\alpha=0,0\leq \alpha\leq C,
\end{array}
\end{equation}
where $\alpha\in\R^m$ is the Lagrangian multiplier vector, $K(\cdot,\cdot)=\langle  \varphi(\cdot),  \varphi(\cdot)\rangle$ is a kernel function \cite{Kernel1}, and $e$ is a vector of ones with an appropriate dimension.
Specifically, $\bfx$ is
classified as $+1$ or $-1$ according to decision
\begin{equation}\label{SVMdecision}
\begin{array}{l}
f(x) =\sum\limits_{i=1}^m\ y_i{K(\bfx,\bfx_i)}\alpha_i +b.
\end{array}
\end{equation}


The above SVM has a nice generalization ability for many problems, but it cannot always classify all of the training samples correctly. Intuitively, for an intelligent model, it should have the ability to remember what it has learned. If we choose a particular kernel such as an RBF kernel function \cite{RBFkernel} with a very small variance, SVM could classify the training samples correctly, but its generalization ability would degenerate largely, which is the over-fitting problem.
Recently, Vapnik and Izmailov \cite{MSVM2021} hired a combination of two RBF kernels into dual problem \eqref{SVMdual} to propose the SVM$^{m}$, which memories the training samples with well generalization. Without proposing the primal problem, SVM$^{m}$ directly solves the following dual problem
\begin{equation}\label{VSVMdual}
\begin{array}{ll}
 \underset{\alpha}{\min}&\frac{1}{2}\alpha^\top\bfY (K_1(\bfX,\bfX)+\tau K_2(\bfX,\bfX))\bfY\alpha-e^\top\alpha\\
\mbox{s.\,t.\,}&  e^\top\bfY\alpha=0,0\leq \alpha\leq C,
\end{array}
\end{equation}
where $K_1(\cdot,\cdot)$ is a generalization RBF kernel with parameter $\sigma$, $K_2(\cdot,\cdot)$ is a memorization RBF kernel with parameter $\sigma^{*}$ ($\sigma^{*}\gg\sigma>0$), and $\tau>0$ is a weighting parameter to balance the generalization and memorization. Its decision becomes to
\begin{equation}\label{SVMmdecision}
\begin{array}{l}
f(x)=\sum\limits_{i=1}^m y_i \alpha_i(K_1(\bfx_i,\bfx)+\tau K_2(\bfx_i,\bfx)) +b.
\end{array}
\end{equation}
Due to the memorization kernel $K_2(\bfx_i,\bfx)$ with very large $\sigma^{*}$, it only affects a small area near $\bfx_i$, and it will not greatly increase the Vapnik-Chervonenkis (VC) dimension \cite{SLT} of the decision.

However, when and why SVM$^m$ works for memorization are implicit, and its geometric interpretations are unclear due to lacking of the primal problem. Thus, it is indistinct to apply the generalization-memorization kernel to other SVM-type models, and it is difficult to apply this kernel to other non-kernel learning models to improve the memorization ability.

\section{Generalization-memorization mechanism}
The goal of the generalization-memorization mechanism is to classify the training data correctly with a well generalization performance. Taking SVM as an example, we give some observations.

\subsection{Memory cost function}
Recalling the SVM's primal problem \eqref{SVMprimal}, for each training sample $(\bfx_{i},y_{i})\in T$, we have
\begin{eqnarray}\label{SVMyf}
\begin{array}{l}
y_{i}f(\bfx_{i})=y_{i}(\bfw^\top\varphi(\bfx_{i})+b)\geq1-\xi_{i}.\\
\end{array}
\end{eqnarray}
To classify all of the training samples correctly, we modify the decision as
\begin{eqnarray}\label{MSVMtraing}
\begin{array}{l}
g(\bfx)=\bfw^\top\varphi(\bfx)+b+\mathcal{I}_{\xi}(\bfx),
\end{array}
\end{eqnarray}
where
\begin{eqnarray}\label{Indicator}
\begin{array}{l}
\mathcal{I}_{\xi}(\bfx)=\left\{\begin{array}{cl}
  y_i\xi_i,&\text{if }\bfx=\bfx_i,\exists \bfx_i\in \bfX,\\
  0,&\text{otherwise}.
\end{array}\right.
\end{array}
\end{eqnarray}
Thus, for each training sample, $y_{i}g(\bfx_{i})\geq1>0$ always holds true with $i=1,\ldots,m$, i.e., all of the training samples are classified correctly by \eqref{MSVMtraing}, which realizes an extremely high memorization ability of SVM. For an unknown sample except it is just the one of the training samples, its decision is the same as the standard SVM. This is an extreme case that SVM memorizes the training samples easily.

Without computing $\xi_i$, decision \eqref{MSVMtraing} can be more general as
\begin{eqnarray}\label{MSSVCW}
\begin{array}{l}
g(\bfx)=f(\bfx)+\mathcal{I}(\bfx),
\end{array}
\end{eqnarray}
where
\begin{eqnarray}\label{Indicator2}
\begin{array}{l}
\mathcal{I}(\bfx)=\left\{\begin{array}{cl}
  y_ic(\bfx_i),&\text{if }\bfx=\bfx_i,\exists \bfx_i\in \bfX,\\
  0,&\text{otherwise.}
\end{array}\right.
\end{array}
\end{eqnarray}
Function $c(\bfx)$ is called memory cost function w.r.t. $\bfx$, where $\bfX$ is its domain of definition and $f(\bfx)$ can be any previous decision function. The larger $c(\bfx)$ , the more cost to memorize $\bfx$. If $c(\bfx_i)$ is large enough, training sample $\bfx_i$ will be classified correctly, e.g., $c(\bfx_{i})\geq \xi_i$ in SVM.

Decision \eqref{MSSVCW} with memory cost function is a realistic path for memorizing training samples. This decision consists of two parts, i.e., the generalization function $f(\bfx)$ and memorization function $\mathcal{I}(\bfx)$.
From this point of view, for machine learning models where empirical risks are concerned, we can hire \eqref{MSSVCW} by setting $c(\bfx_{i})\geq loss(f(\bfx_{i}),y_{i})$. In other words, we can classify all of the training samples correctly by using proper decision \eqref{MSSVCW} no matter whether the corresponding learning model has good memorization ability. At the same time, the generalization of $f(\bfx)$ is retained.

\subsection{Memory influence function}
Though decision \eqref{MSSVCW} could memorize all training samples, there is little impact on generalization. In order to study the influence of memorization on generalization, we construct a new generalization-memorization decision as
\begin{eqnarray}\label{MemorizedSVMdecision}
\begin{array}{l}
g(\bfx)=f(\bfx)+\sum\limits_{i=1}^m y_{i}c(\bfx_{i})\delta(\bfx_i,\bfx),\\
\end{array}
\end{eqnarray}
where $\delta(\bfx_i,\bfx)$ is a function w.r.t. $\bfx$ to measure the influence of memorizing $\bfx_i$, called memory influence function. Generally, the memory influence function could be different. For instance,
supposing the samples that are very similar to $\bfx_{i}$ would be classified into the same class of $\bfx_{i}$, $\delta(\bfx_i,\bfx)$ could be a similarity function between $\bfx_i$ and $\bfx$.
Now, we give some specific memory influence functions, e.g.,
\begin{eqnarray}\label{Influencefunction1}
\begin{array}{ll}
\delta(\bfx_i,\bfx)=\exp(-\sigma ||\bfx_i-\bfx||^{2}),~\sigma>0,
\end{array}
\end{eqnarray}
\begin{eqnarray}\label{Influencefunction3}
\begin{array}{l}
\delta(\bfx_i,\bfx)=\left\{\begin{array}{cl}
  1, &\text{if } ||\bfx-\bfx_i||\leq\varepsilon_i,~\varepsilon_i>0,\\
  0,&\text{otherwise},
\end{array}\right.
\end{array}
\end{eqnarray}
\begin{eqnarray}\label{Influencefunction4}
\begin{array}{ll}
\delta(\bfx_i,\bfx)=\max \{\rho-\|\bfx-\bfx_i\|, 0\}, \rho>0,\\
\end{array}
\end{eqnarray}
or
\begin{eqnarray}\label{Influencefunction2}
\begin{array}{l}
\delta(\bfx_i,\bfx)=\left\{\begin{array}{cl}
  1,& \bfx~\text{is $k$ nearest neighbors of}~\bfx_i,\\
  0,&\text{otherwise}.
\end{array}\right.
\end{array}
\end{eqnarray}
The above functions measure the similarity between $\bfx_i$ and $\bfx$, where the former three functions are symmetric and the last one is not.
Each training sample would have a memory influence on prediction only if it has a nonzero memory cost. The final decision \eqref{MemorizedSVMdecision} is our generalization-memorization decision.

\subsection{Memory model principle}
Based on generalization-memorization decision \eqref{MemorizedSVMdecision}, we discuss the principle of constructing the memory model.
Recalling the traditional SVM, if we use its training model with decision \eqref{MemorizedSVMdecision}, all of the training samples would be classified correctly but the memory cost and memory influence functions should be selected carefully. At this time, the training objective is inconsistent with this decision, which may bring degradation in modeling. As pointed by previous studies \cite{memorization1961,memorizationDNN2018}, the memorized training samples may have positive or negative effects on the model, which may improve or deteriorate its generalization ability. Notice that the memorization ability affects the empirical risk largely, and decision \eqref{MemorizedSVMdecision} can be a very complexity function with a much higher VC dimension \cite{SLT}. We show some results on the expected and empirical risks.

Let $h$ be the VC dimension of $g(\cdot)$ and $G_{gap}=(\ln \left( 2m /h+1 \right) -\ln \eta /4)/(m/h)$ where $\eta\in (0,1)$. Then, expected risk $R(g)$ is bounded by empirical risk $R_{emp}(g)$ and a confidence interval (i.e., generalization gap) from the statistical learning theory (Section 5 in \cite{MSVM2021}) as follows. If $R_{emp}(g)\gg G_{gap}$,
\begin{eqnarray}\label{Risk1}
\begin{array}{l}
R(g)\le R_{emp}(g)+\sqrt{2G_{gap}}
\end{array}
\end{eqnarray}
holds true with probability $1-\eta$,
and if $R_{emp}(g)\ll G_{gap}$,
\begin{eqnarray}\label{Risk2}
\begin{array}{l}
R(g)\le R_{emp}(g)+\sqrt{2}G_{gap}
\end{array}
\end{eqnarray}
holds true with probability $1-\eta$.

Note that the square root of a small number is much larger than that number. Inequality \eqref{Risk1} indicates that confidence interval $\sqrt{2G_{gap}}$ cannot be ignored though empirical risk $R_{emp}(g)$ is much larger than $G_{gap}$. Instead of reducing the VC dimension as much as possible, the previous learning models balance the empirical risk and VC dimension to control the expected risk. However, for a learning model with high memorization ability, it often can achieve zero or almost zero empirical risk. Thus, inequality \eqref{Risk2} indicates that we should reduce the VC dimension as much as possible (i.e., make the model as simple as possible) for a memory model to improve the generalization ability.
Now, a simple and intuitive assumption for memory modeling appears.

\begin{itemize}
\item When a learning machine is built to memorize training samples, we should try not to increase the VC dimension of decision-making and the model complexity.
\end{itemize}

Based on the above principle, we may design different memory cost and memory influence functions to propose the corresponding memory models.

\section{Generalization-Memorization Machines}
By applying the generalization-memorization mechanism into SVM, we build the generalization-memorization machines (GMM) for classification in this section.

\subsection{Hard and soft generalization-memorization machines}
To memorize all of the training samples, we propose a hard generalization-memorization machine (HGMM) under the large margin principle. Our HGMM considers a quadratic programming problem (QPP) as
\begin{equation}\label{Modelstrong}
\begin{array}{l}
\underset{\bfw,b,\bfc}{\min}~~\frac{1}{2}||\bfw||^2+\frac{\lambda}{2}||\bfc||^2\\
 \hbox{s.t.\ }y_{i}(\langle \bfw, \varphi(\bfx_{i}) \rangle +b+\sum\limits_{j=1}^my_jc_j\delta(\bfx_j,\bfx_i))\geq 1,i=1,...,m,\\
\end{array}
\end{equation}
where $\lambda$ is a positive parameter, $\bfc=(c_1,\ldots,c_m)^\top$ denotes the memory costs of the training samples, and $\delta(\bfx_i,\bfx)$ is a user-defined memory influence function. The decision of our HGMM is
\begin{eqnarray}\label{HGMMdecision}
\begin{array}{l}
g(\bfx)=\langle \bfw, \varphi(\bfx) \rangle +b+\sum\limits_{i=1}^my_jc_j\delta(\bfx_i,\bfx).
\end{array}
\end{eqnarray}
Obviously, we hire generalization function $f(\bfx)=\langle \bfw, \varphi(\bfx) \rangle +b$, set the memory cost be variables, and predefine the memory influence function in decision \eqref{MemorizedSVMdecision}. From the constraints of \eqref{Modelstrong}, it requires to memorize all of the training samples. The objective of \eqref{Modelstrong} seeks the large margin with memory costs as lower as possible, and it controls the complexity of the model meanwhile. For the sake of argument, we reformulate problem \eqref{Modelstrong} as
\begin{eqnarray}\label{Model2mat}
\begin{array}{ll}
\underset{\bfw,b,\bfc}{\min}&\frac{1}{2}||\bfw||^2+\frac{\lambda}{2}||\bfc||^2\\
s.t.&\bfY(\varphi(\bfX)^\top\bfw+be+\Delta\bfY\bfc)\geq e,
\end{array}
\end{eqnarray}
where $\Delta\in\R^{m\times m}$ with element $\delta(\bfx_j,\bfx_i)$ ($i,j=1,\ldots,m$). The following theorem holds apparently.
\begin{theorem}\label{exist-1} The empirical risk of HGMM is zero if and only if there is at least a feasible solution to problem \eqref{Model2mat}.
\end{theorem}

The feasibility of problem \eqref{Model2mat} strongly depends on the properties of memory influence matrix $\Delta$. In fact, we have the following sufficient condition for practical applications.
\begin{theorem}\label{thmsufficent2}
The empirical risk of HGMM is zero if $\Delta$ is nonsingular.
\end{theorem}
\begin{IEEEproof}
Consider the feasibility of $\bfY(\varphi(\bfX)^\top\bfw+be)\geq e$. If it is infeasible, we can slack it by adding an extra vector $\xi$, i.e., $\bfY(\varphi(\bfX)^\top\bfw+be)\geq e-\xi$ always holds; Otherwise, set $\xi=0$. Besides, if $\Delta$ is nonsingular, the linear system of equations $\bfY\Delta\bfY\bfc=\xi$ must have a solution, which indicates the feasibility of problem \eqref{Model2mat}. From Theorem \ref{exist-1}, the conclusion holds.
\end{IEEEproof}

Now, we consider the dual problem of \eqref{Modelstrong} via kernel tricks \cite{Kernel1}. Let $\alpha\in\R^m$ be the Lagrangian multiplier vector, the Karush-Kuhn-Tucker (KKT) conditions \cite{KKT} of \eqref{Model2mat} are
\begin{equation}\label{Model1Devfinal}
\begin{array}{ll}
\bfw=\varphi(\bfX)\bfY\alpha,~e^\top\bfY \alpha=0,~\bfc=\frac{1}{\lambda}\bfY\Delta^\top\bfY\alpha,~\text{and}~\alpha\geq0.
\end{array}
\end{equation}
By substituting conditions \eqref{Model1Devfinal} into problem \eqref{Model2mat}, its dual problem is equivalent to
\begin{equation}\label{Model1dual}
\begin{array}{ll}
\underset{\alpha}{\min}&\frac{1}{2}\alpha^\top\bfY(K(\bfX,\bfX)+\frac{1}{\lambda}\Delta\Delta^\top)\bfY\alpha-e^\top\alpha\\
s.t.&e^\top\bfY \alpha=0,\alpha\geq0.
\end{array}
\end{equation}
After solving the above QPP, the final decision is
\begin{equation}\label{Model1Predict}
\begin{array}{l}
g(\bfx)=\sum\limits_{i=1}^my_i\alpha_iK(\bfx_i,\bfx)+b+\sum\limits_{i=1}^my_ic_i\delta(\bfx_i,\bfx).
\end{array}
\end{equation}
In addition, by finding a nonzero component $\alpha_k$ in solution $\alpha$, we have $b=y_k-y_k\sum\limits_{i=1}^my_i(\alpha_iK(\bfx_i,\bfx_k)+c_i\delta(\bfx_i,\bfx_k))$.

When the HGMM is infeasible or memorizing all of the training samples is unnecessary (e.g., data with label noises), our HGMM could be relaxed to a soft generalization-memorization machine (SGMM) by considering a QPP as
\begin{equation}\label{Modelweak}
\begin{array}{ll}
\underset{\bfw,b,\bfc, \eta}{\min} & \frac{1}{2}||\bfw||^2+\frac{1}{2}\lambda||\bfc||^2+ C \sum\limits_{i=1}^m\eta_i\\
 \hbox{s.t.\ } & y_i(\langle \bfw, \varphi(\bfx_{i}) \rangle +b+\sum\limits_{j=1}^my_jc_j\delta(\bfx_i,\bfx_j))\geq 1 -\eta_i,\\
 & \eta_i\geq0,i=1,...,m,
\end{array}
\end{equation}
where $C$ is a positive parameter and $\eta_i$ is a slack variable. Instead of memorizing all of the training samples in the HGMM, our SGMM memorizes the training samples by adjusting $C$, and it approximates to the HGMM when $C$ is large enough. Accordingly, the dual problem of \eqref{Modelweak} is
\begin{equation}\label{Modelweakdual}
\begin{array}{ll}
\underset{\alpha}{\min}&\frac{1}{2}\alpha^\top\bfY(K(\bfX,\bfX)+\frac{1}{\lambda}\Delta\Delta^\top)\bfY\alpha-e^\top\alpha\\
s.t.&e^\top\bfY \alpha=0, 0\leq\alpha\leq C.
\end{array}
\end{equation}
After solving problem \eqref{Modelweakdual} and computing $b$ with $0<\alpha_k<C$ similar to HGMM, our SGMM predicts  samples by \eqref{Model1Predict}.



\begin{figure}[htp]
    \begin{subfigure}{.48\linewidth}\includegraphics[scale=0.19]{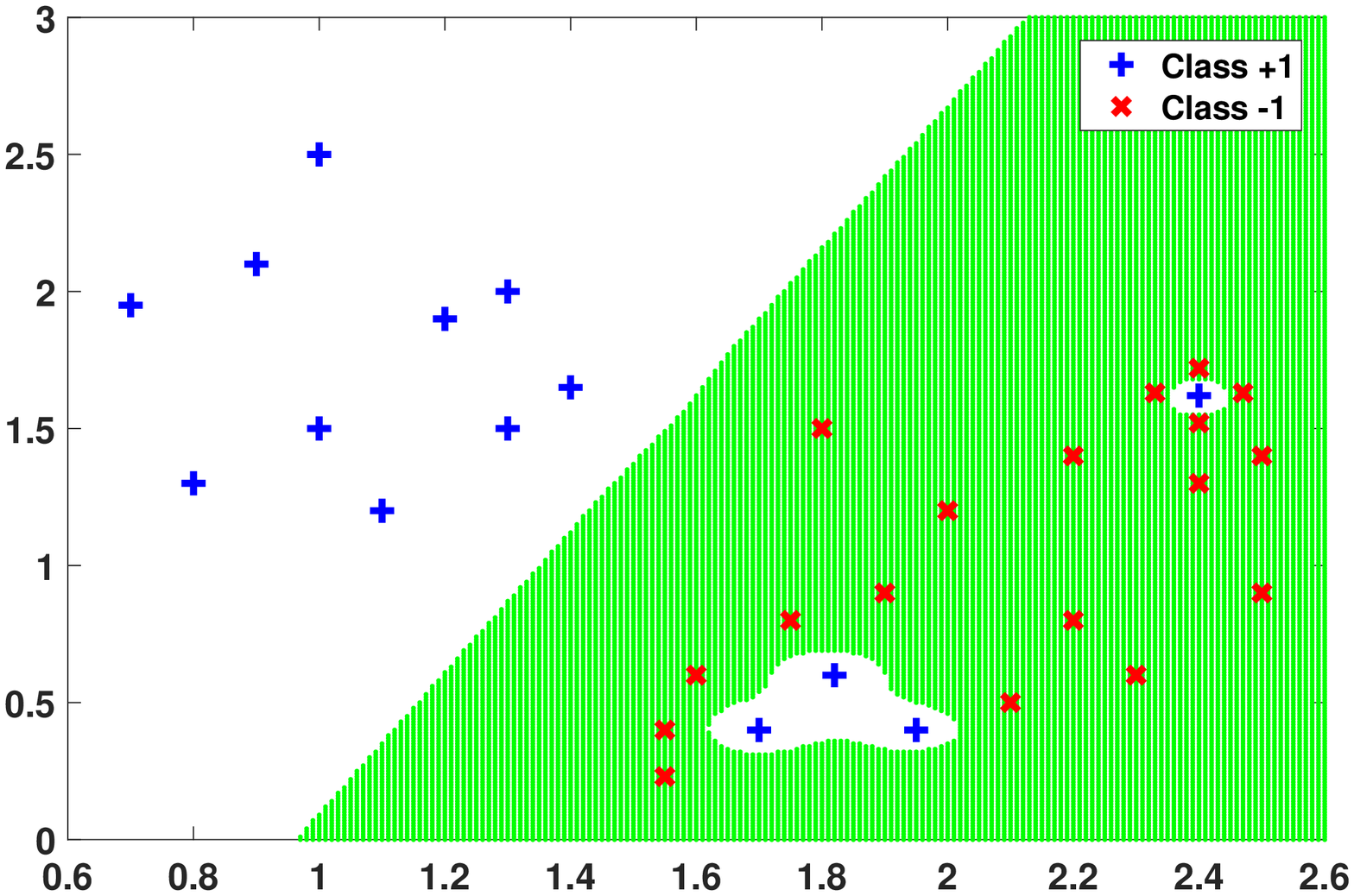}
    \caption{HGMM}
    \end{subfigure}
    \begin{subfigure}{.48\linewidth}\includegraphics[scale=0.19]{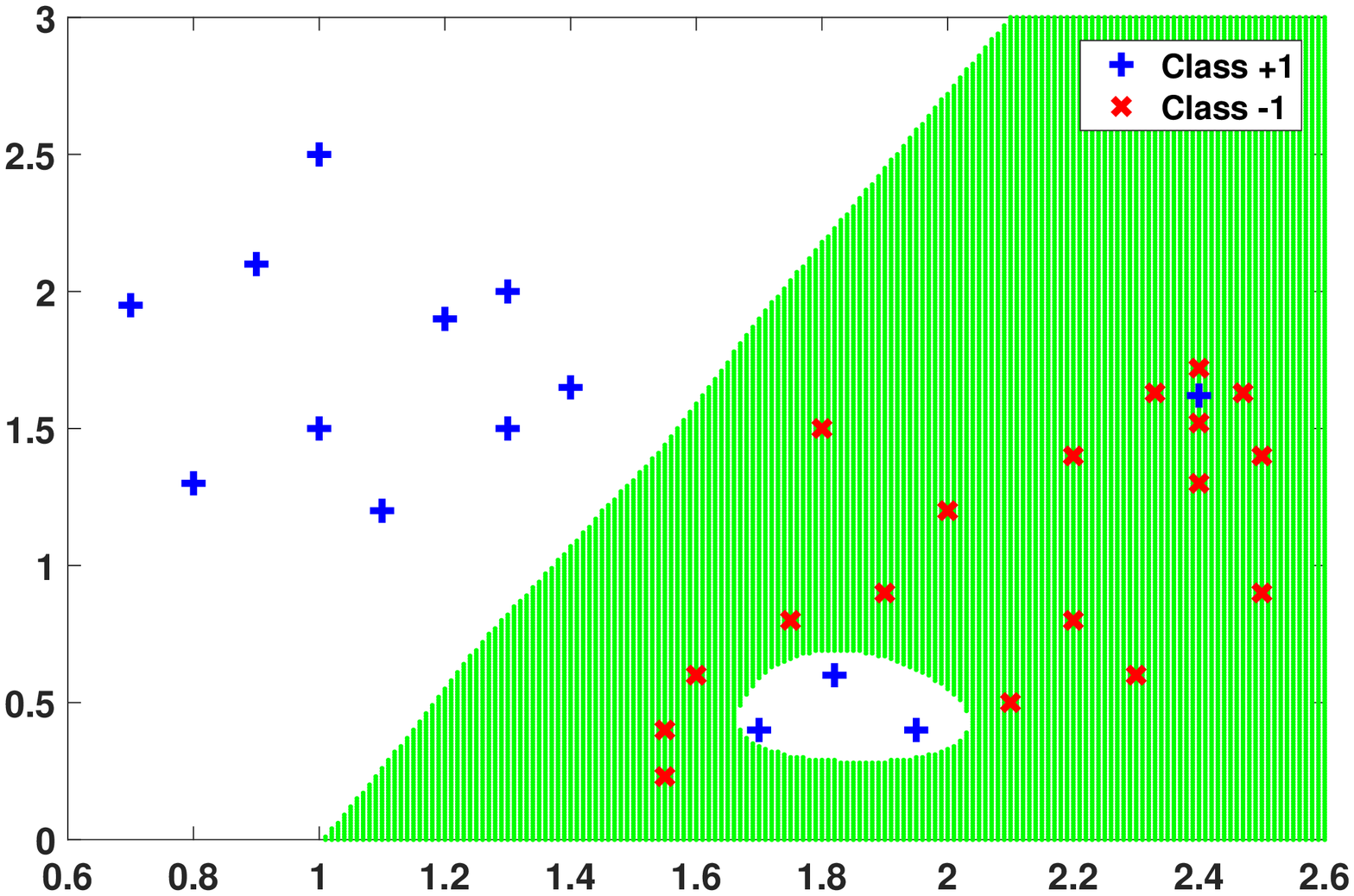}
    \caption{SGMM}
    \end{subfigure}
\caption{A toy example to show the memorization ability of HGMM and SGMM, where memory influence function \eqref{Influencefunction1} and $K(\bfx_i,\bfx_j)=\bfx_i^\top\bfx_j$ are hired in problems \eqref{Model1dual} and \eqref{Modelweakdual}.}
\label{HardSoftMSVM}
\end{figure}

Fig. \ref{HardSoftMSVM} illustrates the memorization abilities of our HGMM and SGMM on a toy example. It can be seen that the HGMM classifies all of the training samples correctly with a smooth boundary similar to the linear SVM, while the SGMM correctly classifies most of the training samples except a positive sample that is surrounded by many negative samples. Thus, our HGMM forces to memorize all of the training samples, and our SGMM has the ability to select and memorize samples that are easy to memorize.

\subsection{Discussion}
In this subsection, let us discuss the relationship between our GMM and SVMs. By analyzing their solutions, we obtain the relationship between our GMM and the classical SVM \cite{SVM95,SVMbook} immediately.
\begin{theorem}\label{SVM=SVMM}
i) If the memory costs of training samples are all zeros after training our HGMM or SGMM, the HGMM is equivalent to hard margin SVM (i.e., problem \eqref{SVMprimal} without slack variables), and the SGMM is equivalent to soft margin SVM (i.e., problem \eqref{SVMprimal}). \\
ii) If the selected $\delta(\bfx_i,\bfx)$ keeps $\Delta$ be the identity matrix,
our HGMM is equivalent to L2 loss SVM with decision \eqref{MemorizedSVMdecision}.
\end{theorem}

For Vapnik's SVM$^m$ \cite{MSVM2021}, we have the following conclusion.
\begin{theorem}\label{SVMm=SGMM}
There exists a memory influence function such that by setting  $K_{2}(\bfX,\bfX)=\Delta\Delta^\top$ and $\frac{1}{\lambda}=\tau$, dual problem \eqref{VSVMdual} of SVM$^m$ is equivalent to dual problem \eqref{Modelweakdual} of our SGMM, and decision \eqref{SVMmdecision} of SVM$^m$ is equivalent to decision \eqref{Model1Predict} of our SGMM.
\end{theorem}
\begin{IEEEproof}
For an RBF kernel $K_2(\cdot,\cdot)$, the kernel matrix $K_2(\bfX,\bfX)$ is symmetric apparently. Thus, we have the real symmetric matrix decomposition $K_2(\bfX,\bfX)=PP^\top$. By selecting the memory influence function $\delta(\cdot,\cdot)$ properly, it is easy to build $P=\Delta$. Then, by setting $\frac{1}{\lambda}=\tau$, problem \eqref{VSVMdual} is equivalent to problem \eqref{Modelweakdual}. On the other hand,
substitute the KKT condition $\bfc=\frac{1}{\lambda}\bfY\Delta^\top\bfY\alpha$ into decision \eqref{Model1Predict}, and let $\frac{1}{\lambda}=\tau$, $K(\bfx,\bfX)=K_1(\bfx,\bfX)$ and $K_2(\bfx,\bfX)=\delta(\bfx,\bfX)\Delta^\top$. We immediately get the conclusion that decision \eqref{Model1Predict} is equivalent to decision \eqref{SVMmdecision}.
\end{IEEEproof}

The above theorem shows that SVM$^m$ is a special case of our SGMM. Moreover, it reveals that primal problem \eqref{Modelweak} of SGMM with a specific memory influence function is actually the primal problem of SVM$^m$. Thus, the clear geometric interpretations of SGMM explain when and why SVM$^m$ works. In the SVM$^m$, RBF kernel matrix $K_2(\bfX,\bfX)$ with large $\sigma^*$ is roughly an identity matrix $I$, and thus, we have its decomposition matrix $\Delta\approx I\approx K_2(\bfX,\bfX)$ (i.e., the memorization term is similar to \eqref{Indicator2}). Subsequently, we have memory costs $\bfc\approx\tau\alpha$. Thus, the larger $\tau$, the higher memorization ability of SVM$^m$, and  memorization kernel $K_2(\cdot,\cdot)$ is actually a memory influence function controls the influence region. If we seek the zero empirical risk by SVM$^m$, a feasible method is to set $C$ and $\sigma^*$ large sufficiently. In fact, the SVM$^m$ approximates to our feasible HGMM in such situation. Finally, we conclude that the memorization ability of SVM$^m$ could be maintained by many other different memorization kernel $K_2(\cdot,\cdot)$ whereas not merely the special RBF kernel.

\section{Experiments}
This section analyzes the performance of our HGMM and SGMM compared with SVMs on several benchmark datasets\footnote[1]{\url{http://archive.ics.uci.edu/ml/index.php}}. Table \ref{Datasets} shows the details of datasets. Thereinto, the classical SVM \cite{SVM95,SVMbook} with the linear or RBF kernel is denoted as SVM$^1$ or SVM$^2$ respectively, the linear generalization kernel and RBF memorization kernel are hired in SVM$^m$ \cite{MSVM2021}, and the linear kernel is hired in our HGMM and SGMM (available at \url{https://github.com/gamer1882/GMM}). All of these models were implemented by MATLAB 2017a on a PC with an Intel Core Duo Processor (double 4.2 GHz) with 32GB RAM, and their QPPs were solved by the same algorithm with the same tolerance. For RBF kernel $K({\bf x}_{i},{\bf x}_{j})=\exp(-\sigma\|{\bf x}_{i}-{\bf x}_{j}\|^2)$, its parameter $\sigma$ was selected from $\{2^{i}|i=-10,-9,\ldots,5\}$, and the other tradeoff parameters of these models were selected from $\{2^{i}|i=-8,-7,\ldots,7\}$.

\begin{table}
\centering
\caption{Details of benchmark datasets}
\begin{tabular}{llrr|llrr}\hline
ID&Name&m&n&ID&Name&m&n\\
\hline
(a)&Bupa&345&6&(k)&Bank&45,211&16\\
(b)&Echocard&131&10&(l)&Creditcard&30,000&23\\
(c)&Heartc &303 &14&(m)&Electrical&10,000&13\\
(d)&Heartstatlog&270 &13&(n)&German&1,000&20\\
(e)&Hepatitis &155 &19&(o)&Htru2&17,898&8\\
(f)&Hourse &300 &26&(p)&Musk&6,598&166\\
(g)&Ionosphere &351 &33&(q)&Qsar&1,055&41\\
(h)&Sonar &208 &60&(r)&Ring&7,400&20\\
(i)&Spect&267&44&(s)&Shoppers&12,330&17\\
(j)&Wpbc &198 &34&(t)&Two&7,400&20\\
\hline
\end{tabular}\label{Datasets}
\end{table}

\begin{table}
\centering
\caption{LOO accuracies (\%) of HGMM with memory influence functions \eqref{Influencefunction1}-\eqref{Influencefunction2} on benchmark datasets}
\begin{tabular}{l|l|lllll}\hline
ID&All&Baseline&\eqref{Influencefunction1} &\eqref{Influencefunction3}& \eqref{Influencefunction4}&\eqref{Influencefunction2} \\
&train&test&test&test&test&test\\\hline
(a)&100.0$\pm$0.00&69.28&$\mathbf{72.46}$&70.43&$\mathbf{72.46}$&69.28\\
(b)&100.0$\pm$0.00&$\mathbf{90.08}$&$\mathbf{90.08}$&$\mathbf{90.08}$&$\mathbf{90.08}$&$\mathbf{90.08}$\\
(c)&100.0$\pm$0.00&$\mathbf{100.0}$&$\mathbf{100.0}$&$\mathbf{100.0}$&$\mathbf{100.0}$&$\mathbf{100.0}$\\
(d)&100.0$\pm$0.00&84.81&$\mathbf{85.19}$&84.81&84.81&84.81\\
(e)&100.0$\pm$0.00&87.10&$\mathbf{87.74}$&87.10&$\mathbf{87.74}$&87.10\\
(f)&100.0$\pm$0.00&81.00&$\mathbf{82.67}$&81.00&81.67&81.00\\
(g)&100.0$\pm$0.00&90.31&95.16&95.16&$\mathbf{96.01}$&90.60\\
(h)&100.0$\pm$0.00&78.37&$\mathbf{87.98}$&81.73&$\mathbf{87.98}$&87.02\\
(i)&100.0$\pm$0.00&82.02&$\mathbf{82.77}$&82.02&82.02&82.02\\
(j)&100.0$\pm$0.00&82.83&$\mathbf{83.33}$&82.83&82.83&82.83\\
\hline
Avg&100.0$\pm$0.00&84.58&$\mathbf{86.74}$&85.52&86.56&85.47\\
\hline
\end{tabular}\label{ResInflu}
\end{table}

Firstly, we test the memorization ability and its influence of our HGMM on several small size datasets. The memory influence functions (i.e., formations \eqref{Influencefunction1}, \eqref{Influencefunction3}, \eqref{Influencefunction4} and \eqref{Influencefunction2}) were preloaded in our HGMM and evaluated by the $m$-fold cross validation (i.e., level-one-out validation, LOO for short). We set the baseline by setting the memory influence function be an identity matrix which is actually L2 loss SVM with decision \eqref{MSVMtraing} according to Theorem 4.3 (ii).
Thereinto, all the parameters $\varepsilon_i$ in \eqref{Influencefunction3} were set to equal to each other and selected from $\{5,1,0.5,0.1,0.05,0.01,0.005,0.001\}$, parameter $\rho$ in \eqref{Influencefunction4} was selected from $\{2^{i}|i=-10,-9,\ldots,10\}$, and parameter $k$ in \eqref{Influencefunction2} was selected from $\{1,2,\ldots,7\}$.
Table \ref{ResInflu} reports their highest LOO training and testing accuracies.
From Table \ref{ResInflu}, it is observed that our HGMM with either memory influence function has $100\%$ training accuracies on all of these datasets. Compare with the training accuracy of baseline, the memory influence function does not directly affect the memorization ability, and thus the memory cost function mainly decides it. Compared with the testing accuracy of baseline, our HGMM with either memory influence function performs better on the LOO testing, which indicates that our generalization-memorization mechanism could improve the learning performance. Among these functions, formation \eqref{Influencefunction1} obtains the highest average LOO testing accuracy, so we hire formation \eqref{Influencefunction1} for our HGMM and SGMM in the rest experiments.

In the following, we compare the memorization and generalization of our HGMM with the SVM$^1$, SVM$^2$ and SVM$^m$ on the small datasets used in Table \ref{ResInflu} by the LOO validation. The highest average LOO training accuracies and their standard deviations together with the corresponding highest LOO testing accuracies were recorded in Table \ref{ResLOO}, and the best testing accuracies for each dataset and average result were bold. It can be seen from Table \ref{ResLOO} that these models except SVM$^1$ can memorize all of the training samples. Thus, the SVM$^2$, SVM$^m$ and our HGMM have a much higher memorization ability than SVM$^1$. Compared to the $100\%$ training accuracies, the testing accuracies do not decrease too much for the SVM$^2$, SVM$^m$ and our HGMM, especially for our HGMM that owns the best LOO testing accuracies on all of these datasets. It indicates that
our HGMM is a suitable path to improve the generalization
ability without over-fitting.
To sum up, our HGMM obtains better performance on both of the memorization and generalization abilities.

\begin{table*}
\begin{center}
\caption{LOO accuracies (\%) on benchmark datasets}
\begin{tabular}{l|llll|llll}\hline
ID&SVM$^1$&SVM$^2$&SVM$^m$&HardMSVM &SVM$^1$&SVM$^2$&SVM$^m$&HGMM\\
&train(\%)&train(\%)&train(\%)&train(\%)&test(\%)&test(\%)&test(\%)&test(\%)\\\hline
(a)&71.50$\pm$0.33 &100.0$\pm$0.00 &100.0$\pm$0.00 & 100.0$\pm$0.00& 69.28&66.09 &72.17 & $\mathbf{72.46}$\\
(b)&93.22$\pm$0.40 &100.0$\pm$0.00 & 100.0$\pm$0.00&100.0$\pm$0.00 &86.26 & 83.97&$\mathbf{90.08}$ &$\mathbf{90.08}$ \\
(c)&100.0$\pm$0.00 & 100.0$\pm$0.00& 100.0$\pm$0.00& 100.0$\pm$0.00&$\mathbf{100.0}$ &$\mathbf{100.0}$ &$\mathbf{100.0}$ &$\mathbf{100.0}$ \\
(d)&86.68$\pm$0.24 & 100.0$\pm$0.00&100.0$\pm$0.00 & 100.0$\pm$0.00&$\mathbf{85.19}$ &78.52 &84.07 &$\mathbf{85.19}$ \\
(e)&90.43$\pm$0.61 & 100.0$\pm$0.00&100.0$\pm$0.00 & 100.0$\pm$0.00&83.23 &85.81 &84.52 &$\mathbf{87.74}$ \\
(f)&81.53$\pm$0.29 & 100.0$\pm$0.00&100.0$\pm$0.00 &100.0$\pm$0.00 &81.33 & 81.33& 81.00&$\mathbf{82.67}$ \\
(g)&95.08$\pm$0.26 &100.0$\pm$0.00 &100.0$\pm$0.00 &100.0$\pm$0.00 &86.89 &$\mathbf{95.16}$ &91.74 & $\mathbf{95.16}$\\
(h)&99.21$\pm$0.35 &100.0$\pm$0.00 &100.0$\pm$0.00 &100.0$\pm$0.00 &74.04 &$\mathbf{87.98}$ &78.37 & $\mathbf{87.98}$\\
(i)&91.02$\pm$0.37 &100.0$\pm$0.00 &100.0$\pm$0.00 & 100.0$\pm$0.00&80.90 &81.27 &81.27 &$\mathbf{82.77}$ \\
(j)&90.86$\pm$0.43 &100.0$\pm$0.00 & 100.0$\pm$0.00& 100.0$\pm$0.00&79.29 &77.78 &82.83 & $\mathbf{83.33}$\\
\hline
Avg&89.95$\pm$0.33&100.0$\pm$0.00&100.0$\pm$0.00&100.0$\pm$0.00&82.64&83.79&84.61&$\mathbf{86.74}$\\
\hline
\end{tabular}\label{ResLOO}
\end{center}
\qquad   \qquad    \qquad   \qquad \quad $^1$ with the linear kernel; $^2$ with the RBF kernel.
\end{table*}

\begin{table*}
\begin{center}
\scriptsize
\caption{Classification accuracies (\%) on grouped benchmark datasets with increasing training samples}
\begin{tabular}{ll|llll|llll}\hline
ID&Training&SVM$^1$&SVM$^2$&SVM$^m$&HGMM&SVM$^1$&SVM$^2$&SVM$^m$&HGMM\\
&number&train&train&train&train&test&test&test&test\\\hline
(k)&50&72.00$\pm$8.03&100.00$\pm$0.00&100.00$\pm$0.00&100.00$\pm$0.00&78.55$\pm$18.63&70.88$\pm$5.48&76.27$\pm$4.16&$\mathbf{79.27}$$\pm$4.35\\
&100&67.05$\pm$7.57&100.00$\pm$0.00&100.00$\pm$0.00&100.00$\pm$0.00&69.00$\pm$15.46&72.17$\pm$3.49&78.35$\pm$3.55&$\mathbf{81.19}$$\pm$2.99\\
&150&66.70$\pm$6.93&100.00$\pm$0.00&100.00$\pm$0.00&100.00$\pm$0.00&77.75$\pm$7.94&71.31$\pm$3.47&79.36$\pm$2.97&$\mathbf{81.57}$$\pm$2.71\\
&200&66.45$\pm$6.32&100.00$\pm$0.00&100.00$\pm$0.00&100.00$\pm$0.00&70.61$\pm$16.99&71.82$\pm$3.07&80.24$\pm$2.54&$\mathbf{81.79}$$\pm$2.42\\
&300&66.05$\pm$5.30&100.00$\pm$0.00&100.00$\pm$0.00&100.00$\pm$0.00&75.01$\pm$9.94&72.72$\pm$2.62&81.22$\pm$1.99&$\mathbf{82.25}$$\pm$1.88\\
&400&65.70$\pm$4.59&100.00$\pm$0.00&100.00$\pm$0.00&100.00$\pm$0.00&76.18$\pm$8.44&70.51$\pm$1.66&81.45$\pm$1.36&$\mathbf{82.42}$$\pm$1.34\\
&500&67.12$\pm$4.57&100.00$\pm$0.00&100.00$\pm$0.00&100.00$\pm$0.00&75.73$\pm$7.02&71.34$\pm$1.32&81.86$\pm$1.28&$\mathbf{82.69}$$\pm$1.25\\
\hline
(l)&50&65.90$\pm$6.70&100.00$\pm$0.00&100.00$\pm$0.00&100.00$\pm$0.00&61.75$\pm$8.68&50.24$\pm$8.08&62.06$\pm$4.62&$\mathbf{63.62}$$\pm$5.94\\
&100&66.10$\pm$4.96&100.00$\pm$0.00&100.00$\pm$0.00&100.00$\pm$0.00&63.15$\pm$13.30&52.26$\pm$7.23&64.15$\pm$2.70&$\mathbf{65.80}$$\pm$4.56\\
&150&64.57$\pm$5.05&100.00$\pm$0.00&100.00$\pm$0.00&100.00$\pm$0.00&66.13$\pm$9.43&52.17$\pm$4.82&65.26$\pm$3.82&$\mathbf{66.50}$$\pm$5.30\\
&200&62.58$\pm$5.28&100.00$\pm$0.00&100.00$\pm$0.00&100.00$\pm$0.00&63.28$\pm$11.67&47.30$\pm$22.79&65.53$\pm$3.31&$\mathbf{66.57}$$\pm$4.45\\
&300&63.53$\pm$3.45&100.00$\pm$0.00&100.00$\pm$0.00&100.00$\pm$0.00&65.72$\pm$12.17&51.05$\pm$25.14&65.95$\pm$2.06&$\mathbf{67.33}$$\pm$3.52\\
&400&62.26$\pm$4.26&100.00$\pm$0.00&100.00$\pm$0.00&100.00$\pm$0.00&59.61$\pm$13.39&50.71$\pm$26.56&67.12$\pm$2.65&$\mathbf{68.16}$$\pm$3.69\\
&500&61.83$\pm$4.29&100.00$\pm$0.00&100.00$\pm$0.00&100.00$\pm$0.00&61.96$\pm$13.94&46.22$\pm$1.77&67.03$\pm$2.35&$\mathbf{67.95}$$\pm$2.91\\
\hline
(m)&50&100.00$\pm$0.00&100.00$\pm$0.00&100.00$\pm$0.00&100.00$\pm$0.00&91.29$\pm$3.10&91.07$\pm$2.30&92.13$\pm$2.89&$\mathbf{92.13}$$\pm$2.88\\
&100&100.00$\pm$0.00&100.00$\pm$0.00&100.00$\pm$0.00&100.00$\pm$0.00&94.60$\pm$1.83&93.57$\pm$1.64&95.44$\pm$1.70&$\mathbf{95.47}$$\pm$1.67\\
&150&100.00$\pm$0.00&100.00$\pm$0.00&100.00$\pm$0.00&100.00$\pm$0.00&96.41$\pm$1.06&94.62$\pm$0.93&96.70$\pm$0.89&$\mathbf{96.71}$$\pm$0.83\\
&200&100.00$\pm$0.00&100.00$\pm$0.00&100.00$\pm$0.00&100.00$\pm$0.00&97.13$\pm$0.83&94.47$\pm$0.65&97.38$\pm$0.60&$\mathbf{97.38}$$\pm$0.63\\
&300&94.98$\pm$16.08&100.00$\pm$0.00&100.00$\pm$0.00&100.00$\pm$0.00&93.27$\pm$15.83&95.20$\pm$0.37&98.22$\pm$0.61&$\mathbf{98.30}$$\pm$0.72\\
&400&79.98$\pm$23.51&100.00$\pm$0.00&100.00$\pm$0.00&100.00$\pm$0.00&78.12$\pm$23.91&95.61$\pm$0.51&98.64$\pm$0.41&$\mathbf{98.73}$$\pm$0.42\\
&500&73.83$\pm$22.70&100.00$\pm$0.00&100.00$\pm$0.00&100.00$\pm$0.00&72.55$\pm$22.54&96.15$\pm$0.44&98.96$\pm$0.38&$\mathbf{99.02}$$\pm$0.35\\
\hline
(n)&50&74.50$\pm$10.30&100.00$\pm$0.00&100.00$\pm$0.00&100.00$\pm$0.00&65.82$\pm$8.71&64.15$\pm$3.54&65.75$\pm$3.55&$\mathbf{67.07}$$\pm$4.27\\
&100&67.35$\pm$12.55&100.00$\pm$0.00&100.00$\pm$0.00&100.00$\pm$0.00&$\mathbf{69.45}$$\pm$13.78&65.15$\pm$3.57&67.56$\pm$2.64&69.05$\pm$3.16\\
&150&64.37$\pm$5.32&100.00$\pm$0.00&100.00$\pm$0.00&100.00$\pm$0.00&62.45$\pm$11.29&65.88$\pm$2.66&68.95$\pm$2.51&$\mathbf{70.30}$$\pm$2.64\\
&200&66.22$\pm$4.66&100.00$\pm$0.00&100.00$\pm$0.00&100.00$\pm$0.00&65.74$\pm$9.32&66.04$\pm$2.41&69.29$\pm$1.93&$\mathbf{70.84}$$\pm$2.18\\
&300&65.67$\pm$3.33&100.00$\pm$0.00&100.00$\pm$0.00&100.00$\pm$0.00&63.41$\pm$8.81&66.49$\pm$2.51&70.39$\pm$1.99&$\mathbf{71.07}$$\pm$1.60\\
&400&65.56$\pm$3.73&100.00$\pm$0.00&100.00$\pm$0.00&100.00$\pm$0.00&66.05$\pm$7.81&66.62$\pm$2.19&70.64$\pm$2.06&$\mathbf{71.33}$$\pm$1.95\\
&500&--&--&--&--&--&--&--&--\\
\hline
(o)&50&82.00$\pm$22.16&100.00$\pm$0.00&100.00$\pm$0.00&100.00$\pm$0.00&79.83$\pm$19.77&84.68$\pm$4.65&95.11$\pm$2.21&$\mathbf{96.75}$$\pm$0.91\\
&100&80.30$\pm$9.94&100.00$\pm$0.00&100.00$\pm$0.00&100.00$\pm$0.00&81.37$\pm$15.74&78.79$\pm$4.11&96.04$\pm$1.50&$\mathbf{96.67}$$\pm$0.87\\
&150&79.87$\pm$13.71&100.00$\pm$0.00&100.00$\pm$0.00&100.00$\pm$0.00&80.68$\pm$23.74&79.31$\pm$2.87&96.43$\pm$0.98&$\mathbf{96.44}$$\pm$0.94\\
&200&80.38$\pm$14.37&100.00$\pm$0.00&100.00$\pm$0.00&100.00$\pm$0.00&85.09$\pm$18.14&81.24$\pm$2.22&$\mathbf{96.84}$$\pm$0.77&$\mathbf{96.84}$$\pm$0.75\\
&300&79.98$\pm$14.82&100.00$\pm$0.00&100.00$\pm$0.00&100.00$\pm$0.00&81.13$\pm$23.50&77.84$\pm$1.92&$\mathbf{96.94}$$\pm$0.56&$\mathbf{96.94}$$\pm$0.55\\
&400&79.91$\pm$9.97&100.00$\pm$0.00&100.00$\pm$0.00&100.00$\pm$0.00&86.95$\pm$11.70&79.16$\pm$2.12&97.04$\pm$0.56&$\mathbf{97.05}$$\pm$0.56\\
&500&76.58$\pm$18.15&100.00$\pm$0.00&100.00$\pm$0.00&100.00$\pm$0.00&79.75$\pm$25.22&80.25$\pm$1.72&97.07$\pm$0.51&$\mathbf{97.08}$$\pm$0.51\\
\hline
(p)&50&100.00$\pm$0.00&100.00$\pm$0.00&100.00$\pm$0.00&100.00$\pm$0.00&69.78$\pm$4.93&$\mathbf{86.98}$$\pm$3.11&71.66$\pm$4.80&84.99$\pm$4.36\\
&100&100.00$\pm$0.00&100.00$\pm$0.00&100.00$\pm$0.00&100.00$\pm$0.00&74.31$\pm$3.98&$\mathbf{91.43}$$\pm$1.27&77.70$\pm$3.38&90.87$\pm$1.93\\
&150&100.00$\pm$0.00&100.00$\pm$0.00&100.00$\pm$0.00&100.00$\pm$0.00&77.88$\pm$3.16&91.79$\pm$1.08&81.74$\pm$3.25&$\mathbf{92.45}$$\pm$1.62\\
&200&100.00$\pm$0.00&100.00$\pm$0.00&100.00$\pm$0.00&100.00$\pm$0.00&80.07$\pm$2.27&$\mathbf{92.42}$$\pm$0.99&84.55$\pm$2.34&92.34$\pm$1.48\\
&300&100.00$\pm$0.00&100.00$\pm$0.00&100.00$\pm$0.00&100.00$\pm$0.00&82.26$\pm$1.69&93.12$\pm$0.89&87.07$\pm$1.50&$\mathbf{93.40}$$\pm$1.53\\
&400&100.00$\pm$0.00&100.00$\pm$0.00&100.00$\pm$0.00&100.00$\pm$0.00&83.33$\pm$1.76&93.95$\pm$0.69&88.93$\pm$1.40&$\mathbf{94.32}$$\pm$0.89\\
&500&100.00$\pm$0.00&100.00$\pm$0.00&100.00$\pm$0.00&100.00$\pm$0.00&84.58$\pm$1.52&94.42$\pm$0.57&90.90$\pm$0.83&$\mathbf{94.99}$$\pm$0.79\\
\hline
(q)&50&100.00$\pm$0.00&100.00$\pm$0.00&100.00$\pm$0.00&100.00$\pm$0.00&73.47$\pm$4.00&$\mathbf{79.96}$$\pm$3.58&78.23$\pm$3.62&79.86$\pm$3.78\\
&100&83.75$\pm$22.03&100.00$\pm$0.00&100.00$\pm$0.00&100.00$\pm$0.00&69.82$\pm$11.24&$\mathbf{82.26}$$\pm$1.99&81.43$\pm$3.25&81.53$\pm$2.21\\
&150&70.03$\pm$8.21&100.00$\pm$0.00&100.00$\pm$0.00&100.00$\pm$0.00&68.09$\pm$10.98&$\mathbf{83.29}$$\pm$1.67&82.52$\pm$2.45&83.22$\pm$2.62\\
&200&70.73$\pm$5.90&100.00$\pm$0.00&100.00$\pm$0.00&100.00$\pm$0.00&69.38$\pm$9.53&$\mathbf{83.98}$$\pm$1.58&83.08$\pm$2.11&83.08$\pm$2.09\\
&300&71.95$\pm$6.49&100.00$\pm$0.00&100.00$\pm$0.00&100.00$\pm$0.00&71.54$\pm$6.76&$\mathbf{84.10}$$\pm$1.59&83.54$\pm$1.88&83.49$\pm$1.89\\
&400&70.92$\pm$6.80&100.00$\pm$0.00&100.00$\pm$0.00&100.00$\pm$0.00&69.08$\pm$8.31&54.66$\pm$28.35&83.86$\pm$1.63&$\mathbf{83.92}$$\pm$1.51\\
&500&--&--&--&--&--&--&--&--\\
\hline
(r)&50&89.10$\pm$17.53&100.00$\pm$0.00&100.00$\pm$0.00&100.00$\pm$0.00&64.61$\pm$5.60&97.77$\pm$0.42&90.25$\pm$3.31&$\mathbf{97.91}$$\pm$0.55\\
&100&70.80$\pm$6.93&100.00$\pm$0.00&100.00$\pm$0.00&100.00$\pm$0.00&66.54$\pm$2.90&98.09$\pm$0.27&95.48$\pm$1.35&$\mathbf{98.20}$$\pm$0.28\\
&150&70.57$\pm$3.17&100.00$\pm$0.00&100.00$\pm$0.00&100.00$\pm$0.00&67.62$\pm$2.41&98.17$\pm$0.22&96.92$\pm$0.96&$\mathbf{98.30}$$\pm$0.19\\
&200&69.38$\pm$3.49&100.00$\pm$0.00&100.00$\pm$0.00&100.00$\pm$0.00&65.91$\pm$1.84&98.26$\pm$0.19&97.62$\pm$0.66&$\mathbf{98.38}$$\pm$0.17\\
&300&69.97$\pm$3.01&100.00$\pm$0.00&100.00$\pm$0.00&100.00$\pm$0.00&68.78$\pm$2.08&98.34$\pm$0.12&98.19$\pm$0.23&$\mathbf{98.41}$$\pm$0.14\\
&400&68.89$\pm$2.88&100.00$\pm$0.00&100.00$\pm$0.00&100.00$\pm$0.00&66.73$\pm$2.84&98.39$\pm$0.09&98.34$\pm$0.12&$\mathbf{98.46}$$\pm$0.16\\
&500&69.03$\pm$2.43&100.00$\pm$0.00&100.00$\pm$0.00&100.00$\pm$0.00&68.13$\pm$1.99&$\mathbf{98.43}$$\pm$0.09&98.41$\pm$0.10&98.42$\pm$0.11\\
\hline
(s)&50&75.20$\pm$19.66&100.00$\pm$0.00&100.00$\pm$0.00&100.00$\pm$0.00&61.79$\pm$15.59&69.11$\pm$6.08&75.22$\pm$6.40&$\mathbf{77.33}$$\pm$5.30\\
&100&64.25$\pm$6.26&100.00$\pm$0.00&100.00$\pm$0.00&100.00$\pm$0.00&62.80$\pm$17.85&71.13$\pm$3.32&$\mathbf{80.88}$$\pm$3.33&80.85$\pm$3.29\\
&150&64.10$\pm$6.70&100.00$\pm$0.00&100.00$\pm$0.00&100.00$\pm$0.00&70.06$\pm$13.86&71.94$\pm$3.59&81.91$\pm$3.42&$\mathbf{81.93}$$\pm$3.26\\
&200&62.15$\pm$6.65&100.00$\pm$0.00&100.00$\pm$0.00&100.00$\pm$0.00&64.92$\pm$18.02&72.26$\pm$3.15&82.03$\pm$4.10&$\mathbf{82.06}$$\pm$3.52\\
&300&61.58$\pm$7.35&100.00$\pm$0.00&100.00$\pm$0.00&100.00$\pm$0.00&66.95$\pm$19.88&50.54$\pm$3.58&$\mathbf{83.08}$$\pm$2.91&83.07$\pm$2.75\\
&400&62.22$\pm$5.10&100.00$\pm$0.00&100.00$\pm$0.00&100.00$\pm$0.00&72.20$\pm$14.44&52.06$\pm$2.66&$\mathbf{83.85}$$\pm$2.39&83.81$\pm$2.47\\
&500&62.45$\pm$4.83&100.00$\pm$0.00&100.00$\pm$0.00&100.00$\pm$0.00&74.57$\pm$14.98&53.47$\pm$2.36&$\mathbf{84.64}$$\pm$1.98&84.57$\pm$2.04\\
\hline
(t)&50&100.00$\pm$0.00&100.00$\pm$0.00&100.00$\pm$0.00&100.00$\pm$0.00&93.86$\pm$2.35&96.29$\pm$1.08&95.86$\pm$0.77&$\mathbf{97.18}$$\pm$0.28\\
&100&100.00$\pm$0.00&100.00$\pm$0.00&100.00$\pm$0.00&100.00$\pm$0.00&94.98$\pm$1.25&96.67$\pm$0.59&96.48$\pm$0.57&$\mathbf{97.39}$$\pm$0.25\\
&150&99.13$\pm$3.88&100.00$\pm$0.00&100.00$\pm$0.00&100.00$\pm$0.00&94.65$\pm$2.46&96.86$\pm$0.53&96.69$\pm$0.42&$\mathbf{97.49}$$\pm$0.17\\
&200&98.32$\pm$5.34&100.00$\pm$0.00&100.00$\pm$0.00&100.00$\pm$0.00&94.32$\pm$3.72&96.97$\pm$0.43&96.89$\pm$0.34&$\mathbf{97.54}$$\pm$0.17\\
&300&98.12$\pm$5.95&100.00$\pm$0.00&100.00$\pm$0.00&100.00$\pm$0.00&94.02$\pm$4.33&97.14$\pm$0.28&97.09$\pm$0.26&$\mathbf{97.60}$$\pm$0.10\\
&400&89.22$\pm$14.16&100.00$\pm$0.00&100.00$\pm$0.00&100.00$\pm$0.00&87.08$\pm$11.31&97.21$\pm$0.19&97.23$\pm$0.22&$\mathbf{97.65}$$\pm$0.10\\
&500&86.28$\pm$3.58&100.00$\pm$0.00&100.00$\pm$0.00&100.00$\pm$0.00&86.33$\pm$3.37&97.30$\pm$0.15&97.35$\pm$0.17&$\mathbf{97.65}$$\pm$0.11\\
\hline
\end{tabular}\label{ResIncrease}
\end{center}
\qquad  \qquad  `--' denotes lack of enough training samples.
\end{table*}

Next, we implemented these models on some larger datasets to further test their generalization abilities with the best empirical risks. For each dataset from (k)-(t) in Table \ref{Datasets}, the number of training samples were selected randomly and increased from $50$ to $500$. We maintained that half of them were from positive class and the other half were from negative class, and the rest consisted of the testing
set. The above procedure was repeated $20$ times to compute the average highest training accuracies and standard deviations together with the corresponding highest average testing accuracies and standard deviations, and the results were reported in Table \ref{ResIncrease}. Obviously, SVM$^1$ cannot get zero empirical risks on most of datasets similar to the results in Table \ref{ResLOO}. Though SVM$^2$ obtains zero empirical risks on all these datasets, its average testing accuracies are around $50\%$ on part of datasets, e.g., on data (l), (q) and (s). The performance of SVM$^2$ on these datasets approximates to the random guess, so it is inferred that the over-fitting problem appears in SVM$^2$. Compared with SVM$^2$, SVM$^m$ and our HGMM do not fall into the over-fitting problem. Among these models, our HGMM maintains the best average testing accuracies with zero empirical risks on most of the comparisons and is comparable with the best ones on the rest several comparisons, which supports the previous conclusions from Table \ref{ResLOO}.

\begin{table*}
\scriptsize
\begin{center}
\caption{Classification accuracies (\%) on grouped benchmark datasets with opposite labels}
\begin{tabular}{ll|lllll|lllll}\hline
ID&FL$^*$&SVM$^1$&SVM$^2$&SVM$^m$&HGMM&SGMM&SVM$^1$&SVM$^2$&SVM$^m$&HGMM&SGMM\\
&\%&train&train&train&train&train&test&test&test&test&test\\\hline
(k)&0\%&64.87$\pm$5.10&64.73$\pm$5.08&64.92$\pm$5.16&100.00$\pm$0.00&74.83$\pm$4.54&78.06$\pm$10.43&$\mathbf{89.00}$$\pm$0.97&88.95$\pm$1.03&82.69$\pm$1.25&87.96$\pm$2.29\\
&5\%&58.70$\pm$6.09&52.56$\pm$6.09&71.67$\pm$1.94&100.00$\pm$0.00&75.28$\pm$4.44&79.40$\pm$16.72&80.02$\pm$23.61&86.96$\pm$1.62&82.54$\pm$1.24&$\mathbf{87.69}$$\pm$2.34\\
&10\%&58.75$\pm$6.40&76.29$\pm$2.68&71.94$\pm$1.67&100.00$\pm$0.00&75.67$\pm$4.34&81.27$\pm$14.33&77.59$\pm$4.40&86.15$\pm$2.15&82.33$\pm$1.24&$\mathbf{87.36}$$\pm$2.49\\
&15\%&62.62$\pm$3.83&76.22$\pm$2.28&74.65$\pm$1.27&100.00$\pm$0.00&84.64$\pm$3.96&75.29$\pm$10.05&77.86$\pm$3.57&84.18$\pm$1.89&81.59$\pm$1.31&$\mathbf{85.32}$$\pm$2.95\\
\hline
(l)&0\%&59.62$\pm$6.20&63.78$\pm$3.75&63.86$\pm$4.41&100.00$\pm$0.00&68.46$\pm$3.77&63.70$\pm$18.00&80.67$\pm$0.60&$\mathbf{80.68}$$\pm$0.55&67.95$\pm$2.91&78.60$\pm$2.41\\
&5\%&58.95$\pm$6.63&54.85$\pm$4.28&70.17$\pm$1.86&100.00$\pm$0.00&68.69$\pm$3.89&65.44$\pm$16.46&76.71$\pm$7.19&$\mathbf{78.64}$$\pm$2.17&67.47$\pm$2.91&78.06$\pm$2.79\\
&10\%&57.36$\pm$5.66&54.54$\pm$4.97&70.00$\pm$1.88&100.00$\pm$0.00&68.50$\pm$4.17&65.63$\pm$17.14&76.79$\pm$6.44&$\mathbf{78.35}$$\pm$2.32&67.08$\pm$3.08&77.58$\pm$3.44\\
&15\%&61.15$\pm$4.46&54.96$\pm$6.33&71.06$\pm$1.91&100.00$\pm$0.00&70.06$\pm$3.31&62.67$\pm$14.57&76.07$\pm$5.14&$\mathbf{77.17}$$\pm$1.50&65.30$\pm$2.94&76.71$\pm$3.01\\
\hline
(m)&0\%&73.83$\pm$22.70&100.00$\pm$0.00&100.00$\pm$0.00&100.00$\pm$0.00&100.00$\pm$0.00&72.55$\pm$22.54&96.15$\pm$0.44&98.96$\pm$0.38&$\mathbf{99.02}$$\pm$0.35&$\mathbf{99.02}$$\pm$0.35\\
&5\%&66.49$\pm$5.51&100.00$\pm$0.00&99.52$\pm$0.26&100.00$\pm$0.00&99.74$\pm$0.11&65.51$\pm$6.23&95.30$\pm$0.49&98.75$\pm$0.28&98.36$\pm$0.39&$\mathbf{98.76}$$\pm$0.32\\
&10\%&68.18$\pm$6.02&100.00$\pm$0.00&99.63$\pm$0.10&100.00$\pm$0.00&99.64$\pm$0.10&65.60$\pm$8.57&95.04$\pm$0.54&98.57$\pm$0.29&98.24$\pm$0.43&$\mathbf{98.67}$$\pm$0.30\\
&15\%&65.95$\pm$5.00&100.00$\pm$0.00&98.97$\pm$0.20&100.00$\pm$0.00&99.06$\pm$0.22&67.08$\pm$5.79&93.85$\pm$0.59&98.31$\pm$0.30&97.92$\pm$0.38&$\mathbf{98.33}$$\pm$0.32\\
\hline
(n)&0\%&56.37$\pm$2.91&100.00$\pm$0.00&62.92$\pm$3.30&100.00$\pm$0.00&62.48$\pm$5.01&78.36$\pm$12.90&81.91$\pm$0.09&81.90$\pm$0.86&76.74$\pm$2.26&$\mathbf{82.05}$$\pm$0.68\\
&5\%&59.71$\pm$5.06&100.00$\pm$0.00&56.67$\pm$1.30&100.00$\pm$0.00&58.43$\pm$3.46&73.61$\pm$16.78&81.91$\pm$0.09&81.83$\pm$0.27&76.25$\pm$2.04&$\mathbf{82.05}$$\pm$0.49\\
&10\%&65.29$\pm$3.41&100.00$\pm$0.00&56.97$\pm$1.65&100.00$\pm$0.00&60.81$\pm$3.69&71.09$\pm$6.17&81.90$\pm$0.09&81.90$\pm$0.31&76.02$\pm$2.24&$\mathbf{82.04}$$\pm$0.78\\
&15\%&60.17$\pm$5.34&100.00$\pm$0.00&57.13$\pm$1.48&100.00$\pm$0.00&63.70$\pm$6.04&71.25$\pm$16.77&81.91$\pm$0.09&81.84$\pm$0.37&76.28$\pm$2.29&$\mathbf{81.92}$$\pm$0.59\\
\hline
(o)&0\%&73.20$\pm$13.74&85.21$\pm$1.30&94.82$\pm$0.80&100.00$\pm$0.00&92.34$\pm$1.65&86.82$\pm$16.72&97.08$\pm$0.09&97.47$\pm$0.32&97.08$\pm$0.51&$\mathbf{97.86}$$\pm$0.27\\
&5\%&68.39$\pm$15.22&82.65$\pm$5.42&94.62$\pm$0.86&100.00$\pm$0.00&92.79$\pm$1.80&88.92$\pm$9.98&94.44$\pm$1.76&97.47$\pm$0.34&97.05$\pm$0.49&$\mathbf{97.86}$$\pm$0.35\\
&10\%&72.02$\pm$13.67&78.53$\pm$4.19&94.41$\pm$0.97&100.00$\pm$0.00&92.21$\pm$1.57&87.10$\pm$17.67&95.93$\pm$1.79&97.45$\pm$0.41&96.99$\pm$0.49&$\mathbf{97.87}$$\pm$0.24\\
&15\%&63.10$\pm$13.95&75.05$\pm$11.66&93.80$\pm$1.06&100.00$\pm$0.00&90.40$\pm$1.81&88.76$\pm$13.49&95.57$\pm$2.15&97.39$\pm$0.45&96.85$\pm$0.48&$\mathbf{97.85}$$\pm$0.24\\
\hline
(p)&0\%&100.00$\pm$0.00&99.77$\pm$0.16&99.98$\pm$0.06&100.00$\pm$0.00&98.30$\pm$1.45&84.58$\pm$1.52&94.43$\pm$0.61&90.92$\pm$0.85&94.99$\pm$0.79&$\mathbf{96.14}$$\pm$0.54\\
&5\%&96.78$\pm$14.40&99.76$\pm$0.17&99.78$\pm$0.20&100.00$\pm$0.00&98.45$\pm$1.35&81.12$\pm$12.83&94.30$\pm$0.61&90.61$\pm$1.03&92.81$\pm$3.28&$\mathbf{96.05}$$\pm$0.51\\
&10\%&86.34$\pm$26.45&99.95$\pm$0.09&99.70$\pm$0.19&100.00$\pm$0.00&98.37$\pm$1.35&72.10$\pm$22.55&94.23$\pm$0.58&90.47$\pm$0.95&87.51$\pm$1.31&$\mathbf{95.99}$$\pm$0.53\\
&15\%&63.31$\pm$6.69&89.13$\pm$1.26&99.37$\pm$0.31&100.00$\pm$0.00&98.68$\pm$1.34&64.97$\pm$12.25&93.92$\pm$0.98&89.42$\pm$1.32&86.75$\pm$1.31&$\mathbf{95.47}$$\pm$0.80\\
\hline
(q)&0\%&66.63$\pm$8.84&80.89$\pm$1.28&96.13$\pm$0.70&100.00$\pm$0.00&84.50$\pm$2.88&74.78$\pm$9.91&86.60$\pm$1.26&86.21$\pm$1.58&84.30$\pm$1.85&$\mathbf{87.70}$$\pm$1.74\\
&5\%&69.03$\pm$6.10&80.99$\pm$1.25&95.89$\pm$0.77&100.00$\pm$0.00&82.93$\pm$3.26&73.25$\pm$8.64&86.48$\pm$1.40&85.87$\pm$1.63&83.64$\pm$1.94&$\mathbf{87.39}$$\pm$1.69\\
&10\%&63.69$\pm$10.72&81.00$\pm$1.14&95.68$\pm$0.87&100.00$\pm$0.00&83.06$\pm$3.15&75.33$\pm$8.06&86.32$\pm$1.52&85.59$\pm$1.66&83.35$\pm$1.70&$\mathbf{87.37}$$\pm$1.58\\
&15\%&65.48$\pm$8.51&80.81$\pm$1.14&95.62$\pm$0.90&100.00$\pm$0.00&81.65$\pm$3.45&75.71$\pm$11.48&86.17$\pm$1.46&85.14$\pm$1.66&82.35$\pm$1.83&$\mathbf{86.89}$$\pm$1.88\\
\hline
(r)&0\%&69.03$\pm$2.43&99.87$\pm$0.15&99.94$\pm$0.09&100.00$\pm$0.00&98.82$\pm$0.43&68.13$\pm$1.99&98.45$\pm$0.08&98.46$\pm$0.10&98.42$\pm$0.11&$\mathbf{98.61}$$\pm$0.06\\
&5\%&69.18$\pm$1.73&99.81$\pm$0.18&99.59$\pm$0.20&100.00$\pm$0.00&98.63$\pm$0.45&67.96$\pm$1.67&98.44$\pm$0.07&98.46$\pm$0.11&98.35$\pm$0.14&$\mathbf{98.62}$$\pm$0.06\\
&10\%&69.28$\pm$1.98&99.76$\pm$0.20&99.41$\pm$0.21&100.00$\pm$0.00&98.49$\pm$0.44&67.95$\pm$1.60&98.44$\pm$0.08&98.46$\pm$0.13&98.06$\pm$0.19&$\mathbf{98.61}$$\pm$0.08\\
&15\%&70.28$\pm$2.14&99.74$\pm$0.18&98.68$\pm$0.26&100.00$\pm$0.00&97.64$\pm$0.44&68.54$\pm$2.40&98.44$\pm$0.09&98.48$\pm$0.15&97.92$\pm$0.18&$\mathbf{98.61}$$\pm$0.07\\
\hline
(s)&0\%&54.78$\pm$5.93&62.46$\pm$3.16&82.84$\pm$2.02&100.00$\pm$0.00&77.55$\pm$5.74&81.03$\pm$10.24&86.70$\pm$0.83&86.89$\pm$1.14&84.57$\pm$2.04&$\mathbf{87.88}$$\pm$1.67\\
&5\%&53.48$\pm$4.03&56.43$\pm$4.09&83.73$\pm$1.71&100.00$\pm$0.00&77.56$\pm$5.87&84.03$\pm$3.77&86.64$\pm$0.44&86.83$\pm$1.13&84.57$\pm$1.92&$\mathbf{88.05}$$\pm$1.31\\
&10\%&58.39$\pm$6.01&54.87$\pm$5.69&82.43$\pm$1.97&100.00$\pm$0.00&73.34$\pm$7.60&82.05$\pm$5.65&86.52$\pm$0.69&86.91$\pm$1.12&84.50$\pm$1.88&$\mathbf{88.29}$$\pm$1.01\\
&15\%&55.14$\pm$4.51&54.36$\pm$5.93&81.22$\pm$1.91&100.00$\pm$0.00&76.01$\pm$5.35&83.80$\pm$3.65&86.43$\pm$0.80&86.99$\pm$1.06&84.67$\pm$1.64&$\mathbf{88.35}$$\pm$0.96\\
\hline
(t)&0\%&86.28$\pm$3.58&98.10$\pm$0.58&98.17$\pm$0.46&100.00$\pm$0.00&100.00$\pm$0.00&86.33$\pm$3.37&97.59$\pm$0.12&$\mathbf{97.69}$$\pm$0.12&97.65$\pm$0.11&97.65$\pm$0.11\\
&5\%&86.44$\pm$4.68&97.17$\pm$0.60&97.91$\pm$0.48&100.00$\pm$0.00&100.00$\pm$0.00&86.48$\pm$4.69&97.32$\pm$0.28&$\mathbf{97.68}$$\pm$0.12&97.66$\pm$0.12&97.66$\pm$0.13\\
&10\%&87.04$\pm$2.97&96.95$\pm$0.58&97.81$\pm$0.44&100.00$\pm$0.00&100.00$\pm$0.00&86.65$\pm$3.19&97.32$\pm$0.29&$\mathbf{97.68}$$\pm$0.11&97.67$\pm$0.10&97.67$\pm$0.10\\
&15\%&88.40$\pm$2.15&96.26$\pm$0.66&96.92$\pm$0.47&100.00$\pm$0.00&99.62$\pm$0.18&88.90$\pm$2.64&97.27$\pm$0.32&$\mathbf{97.66}$$\pm$0.12&97.56$\pm$0.11&97.61$\pm$0.12\\
\hline
\end{tabular}\label{ResChangeY}
\end{center}
\quad `$^*$' denotes fake labels.
\end{table*}

Practically, the zero empirical risk may be unnecessary in some real applications, e.g., the learning tasks with label noises. In the following, we consider the performance of these models compared with our HGMM and SGMM on the benchmark datasets with label noises. For each dataset from (k)-(t) in Table \ref{Datasets}, $500$ training samples were selected randomly as the training set and the rest consisted of the testing set. Thereinto, the labels of $5\%$, $10\%$ or $15\%$ training samples were set to be opposite, and the actual training set was the baseline. These models were implemented on the datasets, and this procedure was repeated $20$ times to record the highest testing accuracies with standard deviations and the corresponding average training accuracies with standard deviations in Table \ref{ResChangeY}. From Table \ref{ResChangeY}, we observe that: i) The training accuracies of these models are not $100\%$ on many comparisons except our HGMM; ii) The testing performances of SVM$^1$ are always lower than other models and unstable with increasing the noises; iii) The testing performances of the other four models decrease regularly with increasing the noises; iv) Their rates of decline are different, where the rates of SVM$^2$ and HGMM are larger and the rate of SVM$^m$ and SGMM are lower; and v) The testing performances of our SGMM are the highest on most of the comparisons and comparable with the highest ones on the rest comparisons.

From the above observations, it is inferred that the performance of SVM$^1$ is unacceptable with lower memorization ability on noises data, whereas the noises strongly affect the generalization ability of models with higher memorization ability such as the SVM$^2$ and HGMM. The eclectic SVM$^m$ and our SGMM perform better than others to avoid the influence of noises as much as possible and preserve the generalization ability. Notice that our SGMM performs better than SVM$^m$ due to it is a special case of SGMM, and we may further improve the performance of SGMM by choosing different memory influence functions. Therefore, for the problem that zero empirical risk is unnecessary, our SGMM is a competitive choice. Additionally, it should be pointed out that SVM$^m$, our HGMM and SGMM hired the linear kernel in the experiments, and their generalization abilities might be improved by choosing other nonlinear kernels for specific problems.


\section{Conclusions}
In this paper, a general generalization-memorization mechanism has been presented by introducing the memory
cost and memory influence functions on decision. Further, applying SVM into this mechanism, two generalization-memorization machines (GMM) have been proposed, where the hard GMM (HGMM) can memorize all of the training samples and the soft GMM (SGMM) could abandon some of them. The optimization problems of GMM are quadratic programming problems similar to that of SVM. Additionally, the recently proposed generalization-memorization kernel and SVM$^m$ are the special cases of our SGMM. Experimental results show the better generalization ability of our HGMM with zero empirical risk and the well adaptation of our SGMM for label noises. In the future work,
it is interesting to use this mechanism into other machine learning problems. Deeper studies on memory cost and memory influence functions are also necessary.

\section*{Acknowledgements}
This work is supported in part by National Natural Science Foundation of
China (Nos. 61966024, 61866010 and 11871183), in part by the Natural Science Foundation of Hainan Province (No.120RC449).

\bibliographystyle{IEEEtran}

\bibliography{FBib}

\begin{thebibliography}{10}
\providecommand{\url}[1]{#1}
\csname url@samestyle\endcsname
\providecommand{\newblock}{\relax}
\providecommand{\bibinfo}[2]{#2}
\providecommand{\BIBentrySTDinterwordspacing}{\spaceskip=0pt\relax}
\providecommand{\BIBentryALTinterwordstretchfactor}{4}
\providecommand{\BIBentryALTinterwordspacing}{\spaceskip=\fontdimen2\font plus
\BIBentryALTinterwordstretchfactor\fontdimen3\font minus
  \fontdimen4\font\relax}
\providecommand{\BIBforeignlanguage}[2]{{%
\expandafter\ifx\csname l@#1\endcsname\relax
\typeout{** WARNING: IEEEtran.bst: No hyphenation pattern has been}%
\typeout{** loaded for the language `#1'. Using the pattern for}%
\typeout{** the default language instead.}%
\else
\language=\csname l@#1\endcsname
\fi
#2}}
\providecommand{\BIBdecl}{\relax}
\BIBdecl

\bibitem{memorizationDNN2018}
S.~Chatterjee, ``Learning and memorization,'' in \emph{International Conference
  on Machine Learning}, 2018, pp. 755--763.

\bibitem{Memory1}
B.~Yang, A.~Ma, and P.~Yuen, ``Revealing task-relevant model memorization for
  source-protected unsupervised domain adaptation,'' \emph{IEEE Transactions on
  Information Forensics and Security}, vol.~17, pp. 716--731, 2022.

\bibitem{DeepL}
L.~Yann, B.~Yoshua, and H.~Geoffrey, ``Deep learning,'' \emph{Nature}, vol.
  521, no. 7553, pp. 436--444, 2015.

\bibitem{MachineL}
M.~Jordan and T.~Mitchell, ``Machine learning: trends, perspectives, and
  prospects,'' \emph{Science}, vol. 349, no. 6245, pp. 255--260, 2015.

\bibitem{memorizationDNN2017}
D.~Arpit, S.~Jastrz{\k{e}}bski, N.~Ballas, D.~Krueger, E.~Bengio, M.~Kanwal,
  T.~Maharaj, A.~Fischer, A.~Courville, and Y.~Bengio, ``A closer look at
  memorization in deep networks,'' in \emph{International Conference on Machine
  Learning}, 2017, pp. 233--242.

\bibitem{memorizationDNN2020}
V.~Feldman, ``Does learning require memorization? a short tale about a long
  tail,'' in \emph{Proceedings of the 52nd Annual ACM SIGACT Symposium on
  Theory of Computing}, 2020, pp. 954--959.

\bibitem{MSVM2021}
V.~Vapnik and R.~Izmailov, ``Reinforced svm method and memorization
  mechanisms,'' \emph{Pattern Recognition}, vol. 119, p. 108018, 2021.

\bibitem{SVM95}
C.~Cortes and V.~Vapnik, ``Support vector networks,'' \emph{Machine Learning},
  vol.~20, pp. 273--297, 1995.

\bibitem{Kernel1}
B.~Sch\"{o}lkopf and A.~Smola, \emph{Learning with kernels}.\hskip 1em plus
  0.5em minus 0.4em\relax Cambridge: MA:MIT Press, 2002.

\bibitem{RandomF}
P.~Resende and A.~Drummond, ``A survey of random forest based methods for
  intrusion detection systems,'' \emph{ACM Computing Surveys (CSUR)}, vol.~51,
  no.~3, pp. 1--36, 2018.

\bibitem{Nonkernel1}
B.~Geng, D.~Tao, C.~Xu, L.~Yang, and X.~Hua, ``Ensemble manifold
  regularization,'' \emph{IEEE Transactions on Pattern Analysis and Machine
  Intelligence}, vol.~34, no.~6, pp. 1227--1233, 2012.

\bibitem{Nonkernel2}
L.~Jing and Y.~Tian, ``Self-supervised visual feature learning with deep neural
  networks: A survey,'' \emph{IEEE Transactions on Pattern Analysis and Machine
  Intelligence}, vol.~43, no.~11, pp. 4037--4058, 2020.

\bibitem{SVMbook}
N.~Deng, Y.~Tian, and C.~Zhang, \emph{Support Vector Machines: Theory,
  Algorithms, and Extensions}.\hskip 1em plus 0.5em minus 0.4em\relax CRC
  Press, Philadelphia, 2012.

\bibitem{RBFkernel}
M.~{\'S}mieja, {\L}.~Struski, J.~Tabor, and M.~Marzec, ``Generalized rbf kernel
  for incomplete data,'' \emph{Knowledge-Based Systems}, vol. 173, pp.
  150--162, 2019.

\bibitem{SLT}
V.~Vapnik, \emph{Statistical Learning Theory}, H.~Simon, Ed.\hskip 1em plus
  0.5em minus 0.4em\relax Wiley-Interscience, New York, USA, 1998.

\bibitem{memorization1961}
R.~Shepard, C.~Hovland, and H.~Jenkins, ``Learning and memorization of
  classifications.'' \emph{Psychological Monographs: General and applied},
  vol.~75, no.~13, p.~1, 1961.

\bibitem{KKT}
R.~Fletcher, \emph{Practical Methods of Optimization}.\hskip 1em plus 0.5em
  minus 0.4em\relax John Wiley and Sons: Chichester and New York, 1987.

\end{thebibliography}

\begin{IEEEbiography}[{\includegraphics[width=1in,height=1.25in,clip,keepaspectratio]{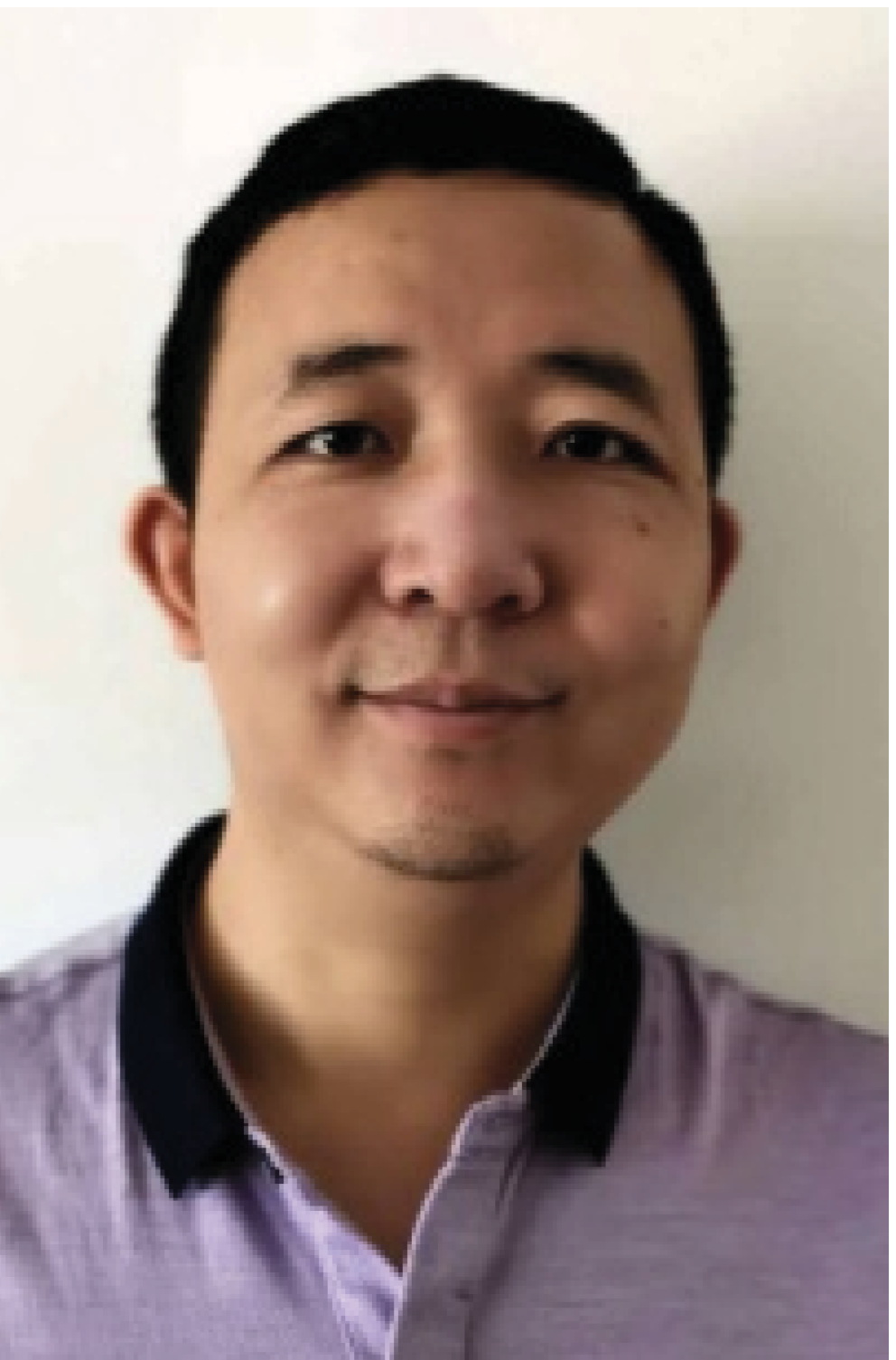}}]{Zhen Wang}
received his bachelor's, master's, and Ph.D. degrees in Mathematics from the Department of Mathematics, Jilin University, Changchun, China, in 2006, 2010, and 2014, respectively. He is currently a Full Professor with the School of Mathematical Sciences, Inner Mongolia University, Hohhot, China. He has published over 20 papers on IEEE TNNLS, IEEE TFS, IEEE TCYB, etc. His research interests include classification techniques, text categorization, and data mining.
\end{IEEEbiography}

\begin{IEEEbiography}[{\includegraphics[width=1in,height=1.25in,clip,keepaspectratio]{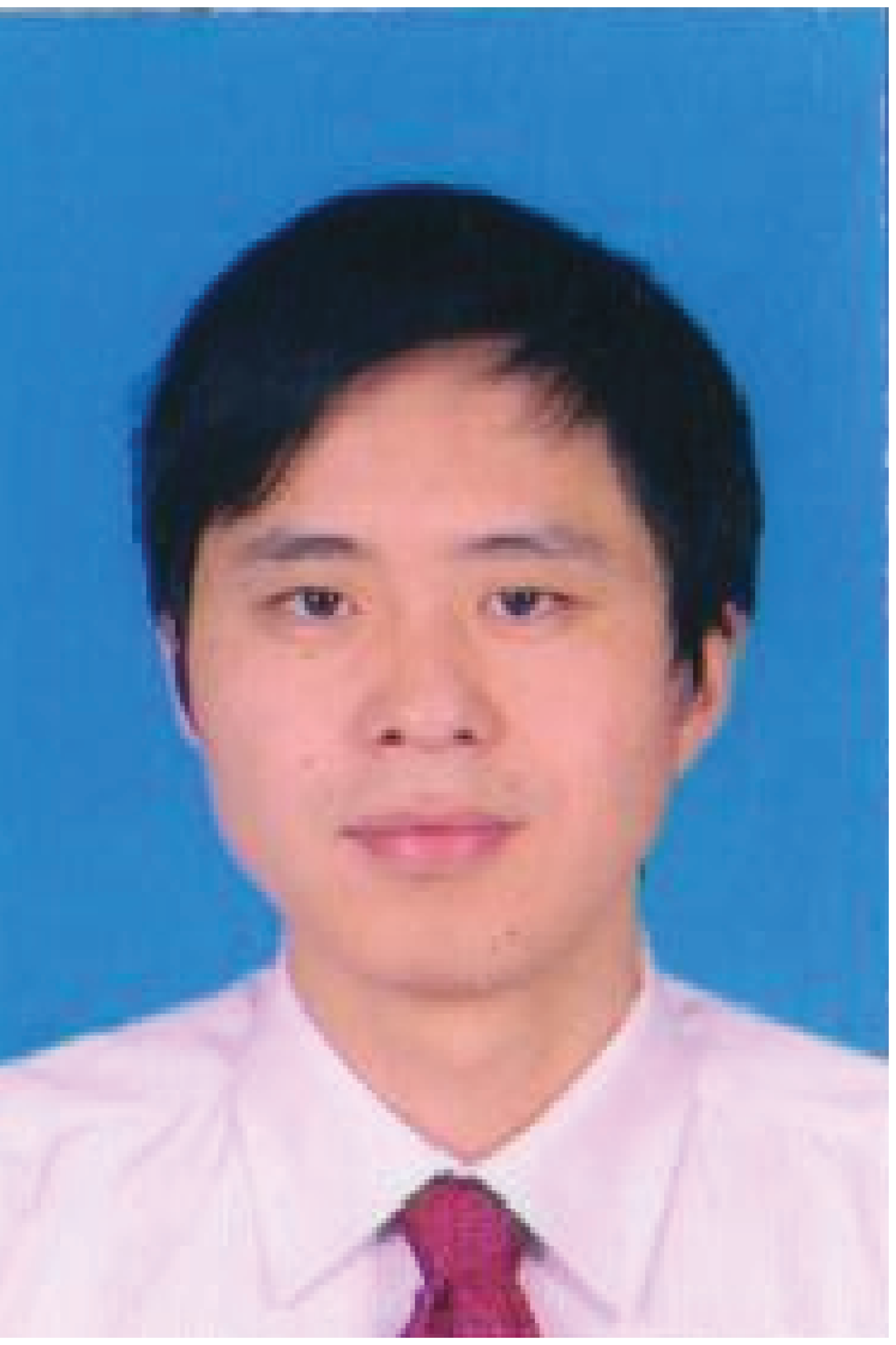}}]{Yuan-Hai Shao}
received his B.S. degree in information and computing science in College of Mathematics from Jilin University, the master's degree in applied mathematics, and Ph.D. degree in operations research and management in College of Science from China Agricultural University, China, in 2006, 2008 and 2011, respectively. Currently, he is a Full Professor at the Management School, Hainan University, Haikou, China. His research interests include support vector machines, optimization methods, machine learning and data mining. He has published over 80 refereed papers on IEEE TPAMI, IEEE TNNLS, IEEE TFC, IEEE TCYB, PR, etc.
\end{IEEEbiography}



\ifCLASSOPTIONcompsoc


\ifCLASSOPTIONcaptionsoff
  \newpage
\fi

\end{document}